\documentclass[
]{ceurart}

\usepackage{listings}
\usepackage{graphicx}
\usepackage{wrapfig}
\usepackage{subcaption}
\usepackage{caption} 
\usepackage{xcolor}
\usepackage{enumitem}
\usepackage{lipsum}
\newlist{requirements}{enumerate}{1}
\setlist[requirements]{label=Req. \arabic*}

\begin{document}

\copyrightyear{2026}
\copyrightclause{Copyright for this paper by its authors.
  Use permitted under Creative Commons License Attribution 4.0
  International (CC BY 4.0).}

\conference{2nd Causal Neuro-Symbolic Artificial Intelligence (Causal Ne{S}y) workshop,
May 10--11 2026, Dubrovnik, Croatia}

\title{ORCA: An End-to-End Interactive Copilot for Optimized Root Cause Analysis}




\author[2]{Phi Nguyen Xuan}
\author[1,3]{Nicholas Tagliapietra}
\author[1]{Lavdim Halilaj}
\author[3,4,5]{Kristian Kersting}
\author[1]{Juergen Luettin}[%
email=Juergen.Luettin@de.bosch.com
]

\address[1]{Robert Bosch GmbH, Germany}
\address[2]{Bosch Global Software Technologies Company Limited, Vietnam}
\address[3]{Computer Science Department, TU Darmstadt, Germany}
\address[4]{Hessian Center for Artificial Intelligence (hessian.AI), Darmstadt }
\address[5]{German Center for Artificial Intelligence (DFKI)}


\newcommand{\marker}[1]{\textcolor{red}{#1}}

\begin{abstract}
  Causal analysis is a crucial task in many domains,
  including manufacturing, social science, and medicine. However, despite recent progress, the conceptual and methodological complexity of causal methods makes them largely inaccessible to domain experts. This gap prevents experts from leveraging these advances and hinders researchers who lack access to real-world data for validation.
  To bridge this divide, we introduce ORCA, a copilot for end-to-end causal analysis.
  ORCA orchestrates agents to understand the user's goals
  and guide them through the most appropriate causal analysis workflow, from fully automatic to highly user-guided execution. 
  It features 
  causal discovery, causal effect estimation, explainability and Root-Cause-Analysis (RCA).
  ORCA evaluates and compares performance, generates key metrics and diagrams, and generates insights through structured reports. We highlight its effectiveness across several real-world use-cases.
\end{abstract}


\maketitle

\section{Introduction}
Causal discovery and inference is a key to scientific understanding and decision making. 
It enables us to move beyond associational patterns and to uncover the underlying mechanisms
that govern observed phenomena \cite{Spirtes2000CausationPredictionSearch}.
By performing a causal analysis, we can reveal how system variables interact and influence outcomes. Therefore, it
plays a crucial role across a wide range of application domains, including 
healthcare \cite{Kellogg2017RCAPatientSafety,Prosperi2020CausalIA, Wu2008RCAJAMA}, 
economics \cite{Imbens2019PotentialOA}, 
health science \cite{Kleinberg2011ARO}, 
genetics \cite{li2025root, Glymour2019ReviewOC},
manufacturing \cite{10.1007/s10845-022-01914-3,Papageorgiou2022RCAReview},
and IT networks \cite{sole2017surveymodelstechniquesrootcause, 10.1145/3501297}.
For example, in medical research we are interested in uncovering the causes of a disease, 
in epidemiology we seek for causal relationships between environmental factors and diseases, 
 and in manufacturing we aim to identify root causes of defects to optimize production.

Despite its importance, modeling causality remains highly challenging, which is why research often stops at the associational level \cite{Kleinberg2011ARO,10.1007/s10845-022-01914-3,Papageorgiou2022RCAReview,sole2017surveymodelstechniquesrootcause}.
In the health domain, for example, although Randomized Control Trials (RTC) 
\cite{cochrane1972effectiveness} are the standard, they are often time-consuming, costly, unethical and/or unfeasible. Within the Pearlian framework, instead, Structural Causal Models \cite{Pearl2009, Spirtes2000CausationPredictionSearch} permit to have a graph representation of the causal mechanisms for the phenomena being modeled, which can often be learned from observational data alone if certain identifiability assumptions are fulfilled. 
But even in the latter case, despite their potential, the usability of such causal models is burdened by high conceptual and methodological complexity. Consequently, real-world applications often remain scattered unless a deep expertise is available. Broader real-world adoption, indeed, is still not achieved.


A typical causal application requires a long and complex and task-dependent pipeline, which includes domain description, data cleaning and preprocessing, causal discovery, effect estimation, and root cause analysis. Each phase requires technical decisions to be impactful, from selecting the appropriate algorithms to tuning hyperparameters and defining evaluation metrics. This intersection of deep methodological expertise and domain-specific knowledge is rarely found in a single practitioner, exacerbating the adoption of advanced causal methods in real-world scenarios.

This adoption gap is further widened by the difficulty of evaluating causal discovery methods. Unlike standard machine learning tasks, causal tasks often lack the standardized benchmarks and evaluation procedures agreed by the community. 
Moreover, the ground-truth causal-graph is rarely available in practice, forcing researchers to rely heavily on synthetic datasets \cite{Cheng2022EvaluationMA} or evaluating on downstream tasks \cite{gentzel2019evaluatinginterventional}. 
Root cause analysis benchmarks are even scarcer \cite{orchard2025root}. Because fully automated solutions cannot be easily validated against a ground truth, effective causal analysis fundamentally requires a human-in-the-loop approach where domain experts iteratively inject structural knowledge and validate assumptions, a process not supported by current tooling.

To bridge the gap between the high demand of causal analysis solutions and its high barrier to entry, we propose ORCA
, an interactive agentic copilot designed to assist end-to-end causal analysis. 
Users can either actively guide the workflow or rely on the copilot's recommendations, removing technical barriers and enabling broader adoption\footnote{A demo video will be made available.}. \paragraph{Contribution:}  Our contributions are summarized as: 


\begin{itemize}
    \item We describe challenges of real-world scenarios and extract a set of requirements derived from these scenarios. 
    \item We present ORCA, our conversational AI-assistant for Causal Analysis developed to address such requirements. We detail our vision, its architecture and further elaborate the main workflow.
    \item We explore real-world use-cases where ORCA can be applied, highlighting its practical advantages and simplicity.
\end{itemize}

\section{Motivating Scenario and Requirements}

Real-world systems exhibit complex causal phenomena that are challenging to model due to their causal structure, high dimensionality and structural mechanisms. 
Consequently, the selection of an optimal modeling pipeline becomes exceedingly challenging for the domain expert. 
We have therefore collected requirements from various scenarios, partially extending those in ~\cite{Saveliev2024CliMBAA}.


\begin{requirements}
    \item \textbf{End-to-End Causal Workflow:} The system must dynamically build the most appropriate causal analysis pipeline, from domain definition, data cleaning, preprocessing, to causal discovery, causal effect estimation, RCA, visualization, and report generation.
    \label{Req: end-to-end-workflow}

    \item \textbf{Intuitive Interaction and Guidance:} ORCA should be friendly to non-experts in causality, it should feature a conversational, natural language interaction via UI. It must actively guide naive users, propose the next steps, and adaptively incorporate user feedback throughout the pipeline.
    \label{Req: Intuitive interaction and guidance}
    \item \textbf{Data Security and Privacy:} Many applications of causality are characterized by highly sensitive data such as clinical or manufacturing data or intellectual properties. The system must therefore ensure strict data-privacy
    and robust role-based access-control. 
    \label{Req: security privacy}  
    \item \textbf{Integration of Domain Knowledge:} To overcome the limitations of observational data, the system must allow to inject of domain knowledge (e.g., prohibited or required causal relationships) sourced from the user, existing documents or LLM-based queries. \label{Req: domain knowledge}
    \item \textbf{Interpretability and Traceability:} The system should provide interpretable results, describe the individual stages in diagrams and allow the user to visualize intermediate results, and clearly explain the reasoning behind its algorithmic choices and root cause identifications.
    \label{Req: interpretability and traceability}

    \item \textbf{Algorithmic Recommendation and Automation:} The system must provide access to state-of-the-art (SOTA) causal methods. It should recommend, starting on AutoML principles, the most appropriate algorithms based on dataset characteristics and user preferences.
    \label{Req: algo recommendation}
\end{requirements}

\section{Related Work}
Traditional causal analysis requires the manual choice of discovery and inference algorithms \cite{Vowels2021DyaLD, Nogueira2022MethodsAT, orchard2025root}, and repeated iterations with domain experts in order to validate results. Recently, however, research started focusing toward automating and assisting these workflows using agentic systems.

\textbf{Causal Reasoning driven by LLMs} Causal Reasoning capabilities on LLMs have been explored in \cite{Zecevic2023CausalPL, Jin2023CanLL}, essentially concluding that their capabilities are likely due to the memorization of cause-effect pairs, and showing how they lack generalization. Nonetheless, \cite{Kcman2023CausalRA, Jiralerspong2024EfficientCG,  Hasan2023OptimizingDC, Liu2024DiscoveryOT, Du2025CD, Ban2025LLMDrivenCD} argue and show their potential in aiding causal discovery tasks by leveraging their internal prior, especially in real-world scenarios where causal assumptions and identifiability fail. Finally, in \cite{vashishtha2025teachingtransformerscausalreasoning, antonucci2023zeroshotcausalgraphextrapolation} authors show the efficacy of LLMs in extracting causal relationships from unstructured textual data.

\textbf{Human-in-the-Loop  and Tool-Instructed Copilots for Data-Science} Copilots for clinical predictive modeling have been introduced in \cite{Saveliev2024CliMBAA} and \cite{saveliev2025humanguideddatacentricllmcopilots}, which was later extended to treatment effect estimation in \cite{berrevoets2025technicalreportfacilitatingadoption}. A first copilot for causal analysis within the Pearlian framework of causality has been proposed in \cite{Wang2025CausalCopilotAA}, which guides the user through causal discovery, causal inference, and causal explanations.
Within the (non-causal) Explainability and RCA domain, \cite{Shan2025RCACT} combines LLMs and granger causality (\cite{Granger1969InvestigatingCR}) for RCA in network incidents,
Further, a neuro-symbolic causal analysis agent for manufacturing has been presented in \cite{Shyalika2025CausalTraceAN, Shyalika2025SmartPilotAC}. 
Finally, an approach for the more general task of causal effect estimation based on an LLM-augmented causal tool has been presented in~\cite{VermaCausalAS}.

In our vision, ORCA extends existing works in interactive and LLM-assisted causal reasoning by filling the gaps and integrating (1) SOTA causal discovery and RCA methods, (2) automated method recommendation, (3) scalable to use cases with partial causal information and/or without a causal graph, (4) integration of domain knowledge from both structured and unstructured sources.

\begin{figure}
  \centering
  \includegraphics[width=0.98\linewidth]{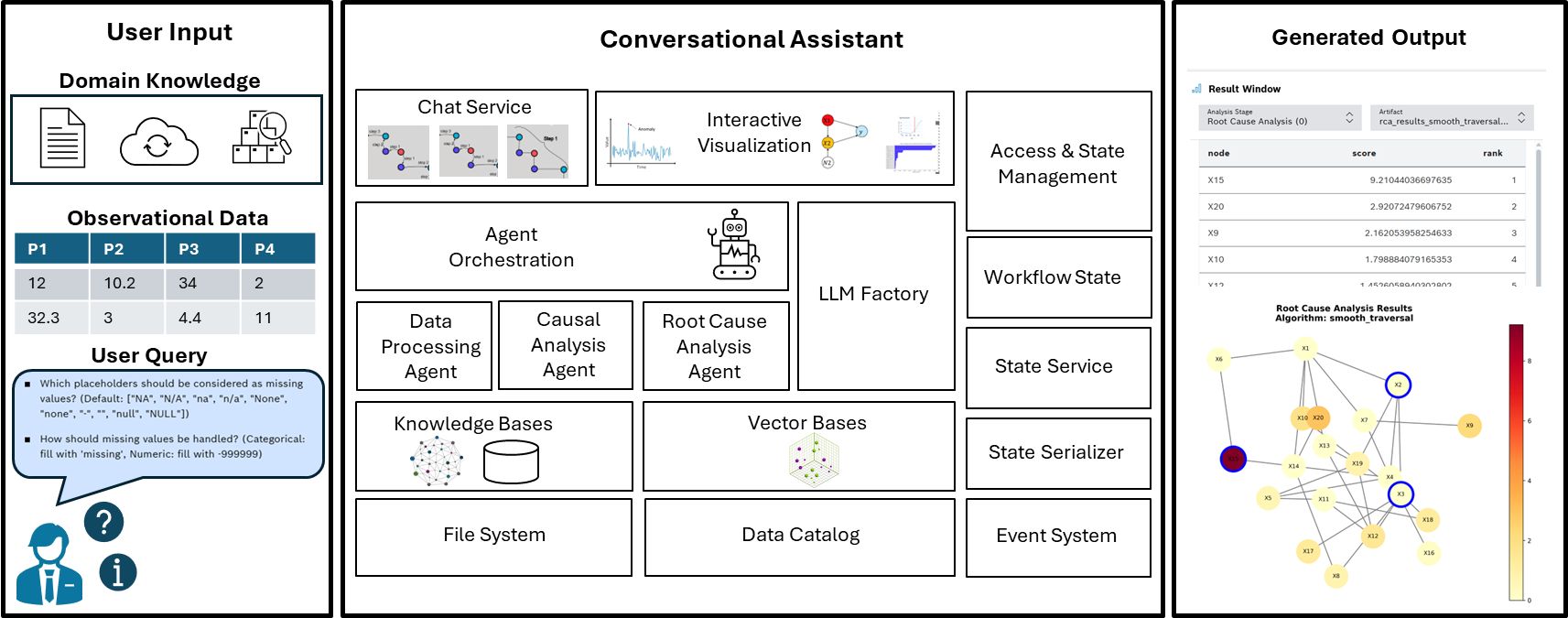}
  \caption{ORCA is an assistant for Causal Analysis. By interacting with the user and its provided data and information (left), it orchestrates the execution of the most appropriate workflow for a specific causal analysis task. It features a wide variety of methods (center) 
  and generates reports tailored to the user needs (right).
  }
    \label{fig:concept_architecture}
\end{figure} 

\section{ORCA}

We designed ORCA to address the outlined requirements. A conceptual overview is shown in Fig.~\ref{fig:concept_architecture} comprising there main pillars: 1) User Input - offering the possibilities to retrieve domain knowledge, observational data as well as given queries; 2) Conversational Assistant - built on top of a multi-agent framework to streamline the interaction across various integrated modules; and 3) Generated Output - responsible for providing the results in various formats and visualization options. 

\subsection{System Architecture}
\begin{figure}
  \centering
  \includegraphics[width=0.98\linewidth]{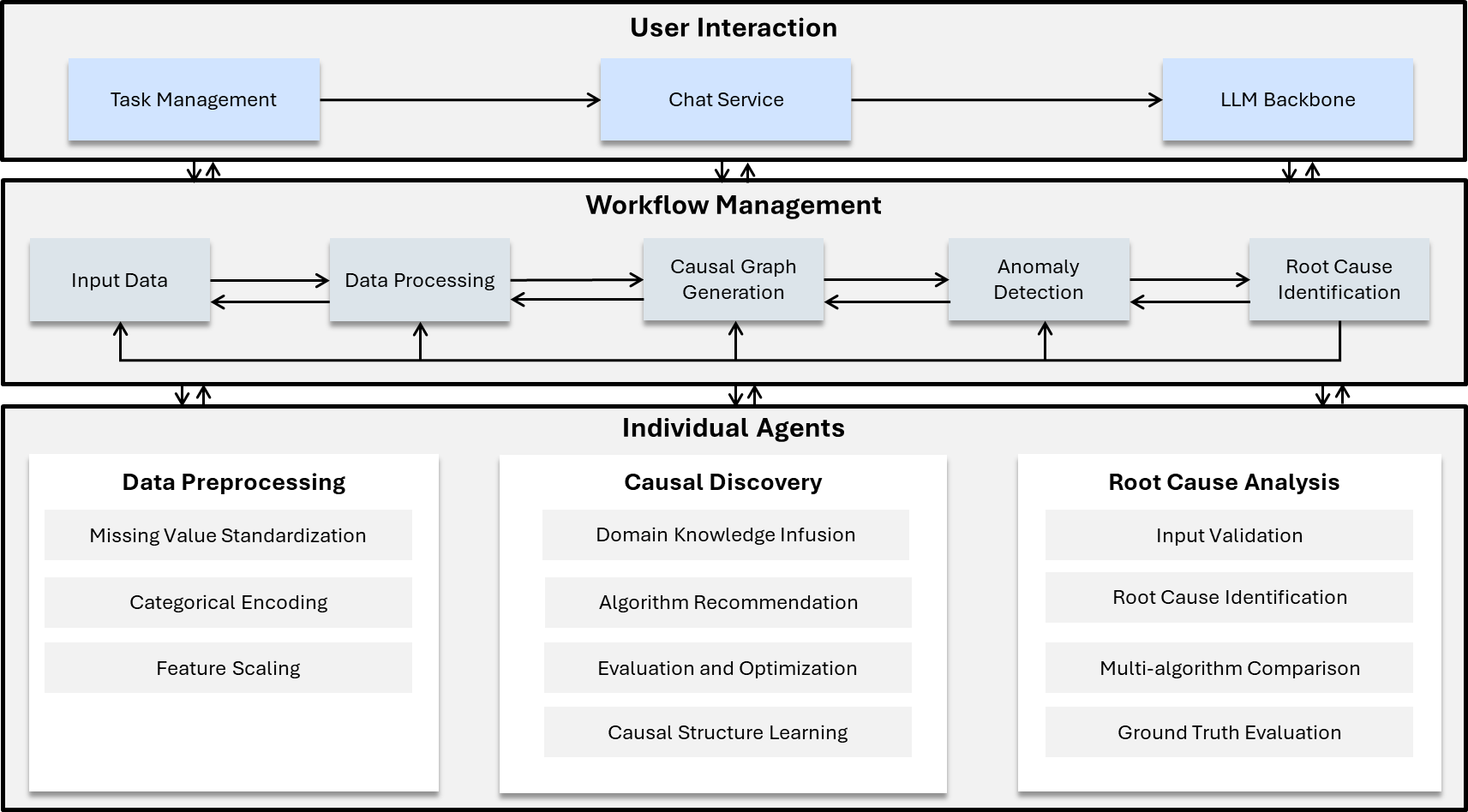}
  \caption{Multi-agent architecture illustrating the workflow management backbone (middle) that orchestrates the whole workflow with user interaction (top) and executing required individual agents (bottom). }
    \label{fig:interactive_workflow}
\end{figure}

The architecture of ORCA is designed to provide a robust, no-code environment for causal analysis, directly addressing the needs of domain experts. 
As illustrated in Fig.~\ref{fig:interactive_workflow}, the framework can be divided into (1) User Interaction, that takes care of task management, chat service and LLM backbone, (2) Workflow Management, responsible for the whole workflow chain with user interaction and execution of Individual Agents (3). 

\paragraph{Request Management Unit (\ref{Req: end-to-end-workflow}).}
ORCA manages workflow states via a centralized orchestration mechanism. 
The system routes API calls to remotely hosted LLMs while offloading intensive algorithmic workloads to available GPU clusters. A central Generation Unit dynamically translates the orchestrator's high-level reasoning into executable Python scripts, securely bridging the gap between natural language intents and the underlying state-of-the-art (SOTA) causal libraries.

\paragraph{Computation Time Estimation (\ref{Req: Intuitive interaction and guidance}).}
Algorithms involving causal models are in large part intractable. Especially causal discovery and complex RCA algorithms scale poorly with the dimensionality of the data, resulting in explosion of processing times. 
To mitigate this critical operational constraint, the system evaluates the selected method's theoretical complexity, the dataset's dimensionality, and the currently available hardware (e.g., GPU clusters). 
It then provides a runtime estimate that can guide users to make informed decisions between accuracy and execution speed. The motivation is that distinct real-world applications might have different time constraint, highlighting a critical trade-off: Sometimes a diagnosis or an anomaly in a manufacturing line have to be carried on very short timelines. Other times, instead, a longer compute time is an acceptable price for higher accuracy. 

\paragraph{Security and Privacy (\ref{Req: security privacy})} To ensure secure handling of data, the system uses TLS-encrypted communication between the user interface and backend services via authenticated APIs. Access to datasets and outputs from Causal Discovery or Root Cause Analysis is controlled through role-based access control (RBAC) integrated with the authentication and authorization layer. Each interaction runs within an isolated session context to prevent cross-user data leakage.

\subsection{User Interaction}
Causal models have been historically built by bridging data with human expertise. To facilitate this, ORCA offers interaction in two modalities: Natural language and Graphical User Interface (GUI).

The core of the interaction happens via a Natural Language Interface (NLI). In the NLI, the user can state the goal, steer ORCA, and inject prior knowledge. Viceversa, the system can provide suggestions on the optimal pipeline and recommend algorithms.
To further facilitate the interaction, we complement with a GUI that permits to have more granular control and inspect results more easily. 

\paragraph{Knowledge and Data Ingestion (\ref{Req: domain knowledge})} Each session starts with the user providing data and the relevant context in natural language. Users can upload observational/interventional datasets along with domain knowledge in varied forms (e.g.\, manuals, technical reports, knowledge graphs, databases). While the provided data is essential for the causal algorithms, the agents can use the additional domain knowledge to extract as much information as possible. For example, this prior knowledge can be used to clarify the meaning of variables and remove ambiguity, or to extract specific causal relationships, or to absorb knowledge about diagnoses or anomalies that happened in the past.

\paragraph{Interactive Orchestration (\ref{Req: Intuitive interaction and guidance})} Since causal inference based on purely observational data is rarely possible or statistically significant enough, the system strongly relies on prior knowledge and user validation. In practice, ORCA employs a state management module that logs each step that has been performed along with its results. This permits to track the progress in relation to the underlying objective defined by the user. Consequently, in the user disagrees with the current pipeline, it can navigate back to previous steps to, for example, integrate new prior knowledge or execute a different algorithm without losing its global context. The LLM continuously uses the session history to ensure that each reasoning step is logically consistent with past actions.

\paragraph{Data Visualization and Report Generation (\ref{Req: interpretability and traceability})} Along the whole execution, the system offers the possibility to explore and visualize the data including descriptive statistics and exploratory data analysis (EDA). Finally, ORCA automatically synthesizes the whole causal analysis and uncovered causal phenomena into a report. It compiles each decision, causal graph, and downstream method into a comprehensive and human-readable report. This final synthesis aims to remove complex algorithmic outputs and make the causal analysis accessible, reproducible, and interpretable to a broader audience.

\subsection{Concrete Features}
\paragraph{SOTA Algorithms and Recommendations (\ref{Req: algo recommendation}).}
ORCA provides a comprehensive suite of SOTA tools and methods covering the entire causal analysis workflow, which we list in Table \ref{tab:orca_features}. 
ORCA automatically analyzes dataset characteristics 
to recommend the most appropriate algorithms. Once selected, ORCA applies AutoML principles to search for the optimal method and associated hyperparameters to the specific task.
To ensure rigorous evaluation and reproducibility, the system also integrates a suite of standard metrics and off-the-shelf benchmarking datasets. 

\paragraph{Datasets and Benchmarks} To facilitate empirical evaluation, establish empirical baselines, and ensure reproducibility, ORCA natively implements a suite of standard benchmark datasets and data generation simulators. Real-world and benchmark datasets include \textit{CausalChambers} \cite{gamella2025chamber}, \textit{CausalMan} \cite{tagliapietra2025causalmanphysicsbasedsimulatorlargescale}, \textit{Petshop} \cite{hardt2024petshop}, \textit{Sockshop} \cite{Ikram2022RootCA}, and \textit{ProRCA} \cite{dawoud2025prorcacausalpythonpackage}. Furthermore, the system features simulations for random graphs, allowing users to generate Erdős–Rényi \cite{erdos59a} and Scale-Free \cite{albertbarabasi1999} graphs of arbitrary size and functional form (linear or non-linear). These simulators support additive noise drawn from Gaussian, Gumbel, or Uniform distributions. To rigorously benchmark Root Cause Analysis and causal explainability, ORCA can systematically synthesize root causes by injecting hard, soft, single, or multiple interventions into the simulated data.

\begin{table}[t]
    \refstepcounter{table}

    \noindent\textbf{Table \thetable.} Overview of ORCA Supported Features, Algorithms, and Metrics.
    
\small
\begin{tabular}{@{}p{0.12\linewidth} p{0.67\linewidth} p{0.16\linewidth}@{}}
\toprule
\textbf{Module} & \textbf{Supported Methods \& Features} & \textbf{Metrics / Output} \\ \midrule

\textbf{Preprocessing \& Data Analysis} & 
\textbf{Data Cleaning \& Preprocessing:} Data type conformance checking, categorical data encoding. Missing value detection, imputation, and dropping of sparse parameters/samples. Unique value, parameter, and sample treatment. Normalization (zero-mean unit-variance, robust, min-max scalers). \newline
\textbf{Data Analysis:} Descriptive statistics, bivariate and multivariate Exploratory Data Analysis (EDA), and data distributions. & 
Cleaned and encoded datasets, descriptive statistics, visualization plots. \\ \midrule

\textbf{Causal Discovery}\footnotemark[1] & 
\textbf{Static Data:} \newline
\textit{Constraint-based:} PC \cite{Spirtes2000CausationPredictionSearch}, FCI \cite{Spirtes1995CausalII}. \newline
\textit{Score-based:} GES \cite{Chickering2002OptimalSI}, XGES \cite{nazaret2021extremely}, GRaSP \cite{Lam2022GreedyRO}. \newline
\textit{Continuous:} NOTEARS \cite{Zheng2028Notears}, GOLEM \cite{ng2021rolesparsitydagconstraints}, CORL \cite{Wang2021OrderingBasedCD}. \newline
\textit{Functional/Hybrid:} LiNGAM \cite{shimizu2006Lingam}, ANM \cite{Hoyer2008NonlinearCD}, PNL \cite{Zhang2009OnTI}. \newline
\textit{LLM-based:} CausalSteward \cite{tagliapietra2026causalsteward}. \vspace{1.5mm} \newline
\textbf{Time Series Data:} \newline
\textit{Constraint-based:} PCMCI \cite{Runge2017DetectingAQ}. \newline
\textit{Granger Causality:} NeuralGC \cite{Tank2021NeuralGC}. & 
Structural Hamming Distance (SHD), normalized SHD. \\ \midrule

\textbf{Root Cause Analysis (RCA)}\footnotemark[2] & 
\textbf{Graph-Required:} Traversal \cite{liu2021microhecl}, CI-RCA \cite{Li2022CausalIR}, Counterfactual attribution \cite{budhathoki2022causal}, Score Traversal \cite{orchard2025root}. \newline
\textbf{Graph-Free:} RCD \cite{Ikram2022RootCA}, Cholesky Composition \cite{li2025root}, Score Ordering \cite{orchard2025root}. & 
Precision, Recall, F1, Accuracy, NDCG, MRR, MAP@k. \\ \bottomrule
\label{tab:orca_features}
\end{tabular}

\end{table}
\footnotetext[1]{Includes algorithms derived from the gcastle library (https://github.com/huawei-noah/trustworthyAI).}
\footnotetext[2]{Includes algorithms derived from the RCA library (https://github.com/amazon-science/RCAWithMissingStructuralKnowledgeCode).}

\section{Case Studies}

To highlight the versatility of ORCA's impact and showcase its intuitive usage, we present common problem statements on different domains and describe how ORCA guides the user through a full solution strategy, as well as providing the respective algorithm outcomes and visualizations of the strategy. 
\begin{enumerate}
    \item \textbf{Cloud Computing Provider} 
    A cloud service provider operates complex applications composed of numerous microservices. The system experiences an outage or a significant performance degradation, triggering alerts for multiple services. These could be as example infrastructure failures, application-level failures or external dependency failures.  The flood of alerts makes it difficult for the site reliability engineering team to distinguish between symptoms and actual root causes. The provider wants to detect the root cause of the failure, which is the specific service or component that initiated the problem. ORCA can rapidly and automatically identify the true root cause, pinpointing to the precise point of failure to resolve the issue and minimize down time.
    
        \item \textbf{Retail:} 
        A clothing company is facing declining overall profitability due to rising material costs and shifting consumer behavior. To counteract this trend, they need to increase their profit margins. Upon analysis, they discover a significant and unexplained variance in profit margins across different product categories  and sales channels. The leadership team needs to understand the causal drivers of this variance to devise an effective strategy. The company's data analytics team has access to a massive dataset containing sales transactions, product costs, pricing information, and promotional discount data. While they can see correlations, they cannot distinguish between cause and effect. ORCA uses causal discovery to model the relationships between their key business levers. Then it can be used to estimate the causal impact, e.g. how does a change in the discount-rate effect its profit-margin. Validating different factors with this  model, the company can decide on the most effective measures to maximize their profit.
    \item \textbf{Manufacturing:} A manufacturing company assembling magnetic valves and hydraulic blocks to hydraulic units is detecting increased leakage failures during pressure testing, leading to production yield decrease. The process engineers are interested to find the cause of this leakage. 
    They consider partially known functional relationships in the system and let ORCA perform the Cholesky Composition on their data to get a ranked list of potential root-causes. 
    Once tested for the top three provided estimates, they find the surface roughness variation in hydraulic blocks was causing seals to not fully conform to the surface, ultimately leading to the defects. 
\item \textbf{Semiconductor Fabrication:} 
    A semiconductor fabrication plant (fab) manufactures complex integrated circuits on silicon wafers. The production process is incredibly intricate, involving hundreds of discrete steps (like etching, lithography, deposition, and cleaning) that take place over several months. Each production step is meticulously monitored, generating thousands of process parameters and sensor readings that are collected and stored. The fab experiences a "yield excursion"—a sudden, unacceptable drop in the percentage of functional chips on its wafers. 
    Process engineers are now face with the task to find the root cause of the defects.
    In this scenario, a swift response is critical to prevent high losses from scrapped wafers. ORCA analyses the dataset to automatically detect the root cause of the yield drop. The goal is to pinpoint the specific event or parameter drift that initiated the failure cascade. For example, identifying a subtle, out-of-spec deviation that is causally linked to the defect. For example, discovering that the plasma pressure in a specific etching tool was 0.2\% below its target range for a 30-minute window two months ago.  By rapidly and automatically identifying the true root cause, the fab can immediately implement corrective actions to protect production yield.
 
\end{enumerate}

\begin{figure*}[tb]

   \begin{subfigure}[b]{0.49\textwidth}
        \centering
        \includegraphics[height=5.2cm]{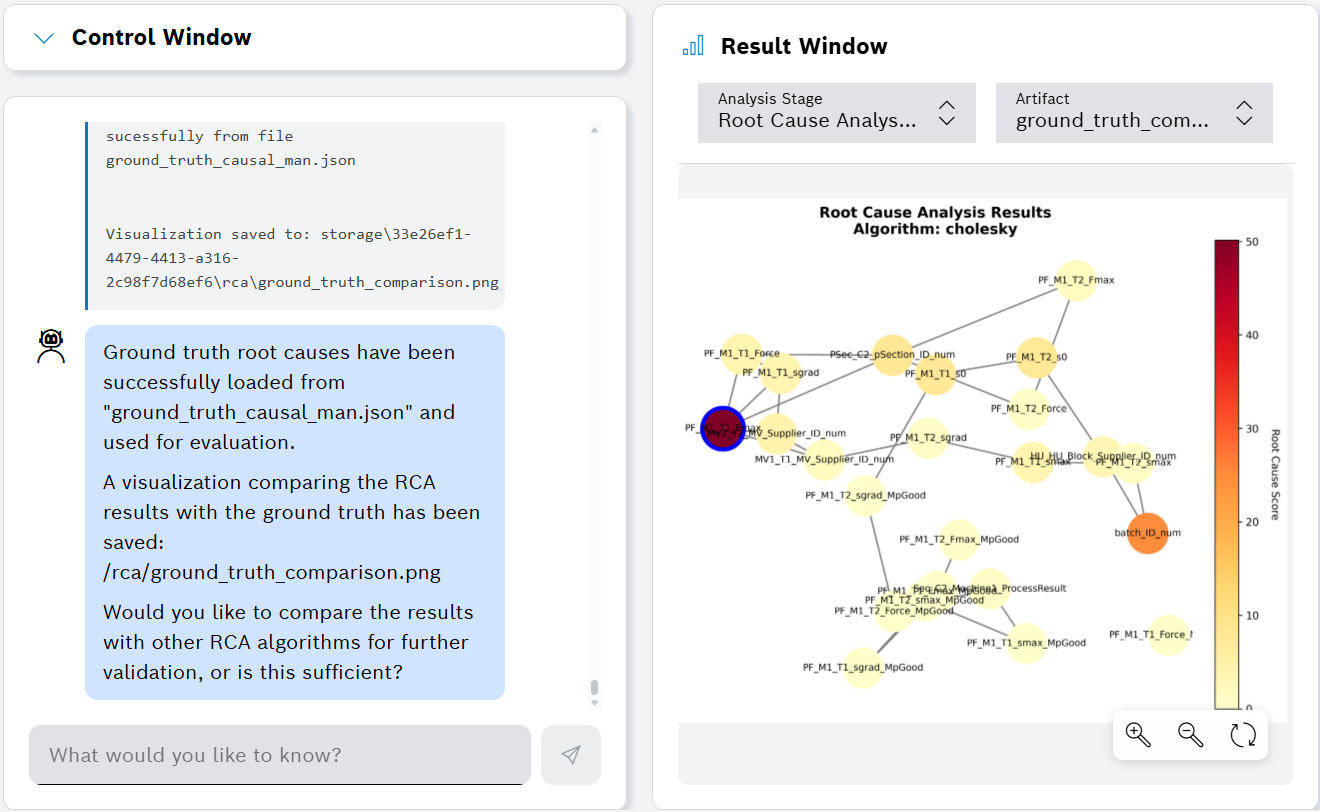}
        \caption{\small Manufacturing}
        \label{fig:example_1}
    \end{subfigure}
    \hfill
    \begin{subfigure}[b]{0.49\textwidth}
        \centering
        \includegraphics[height=5.2cm]{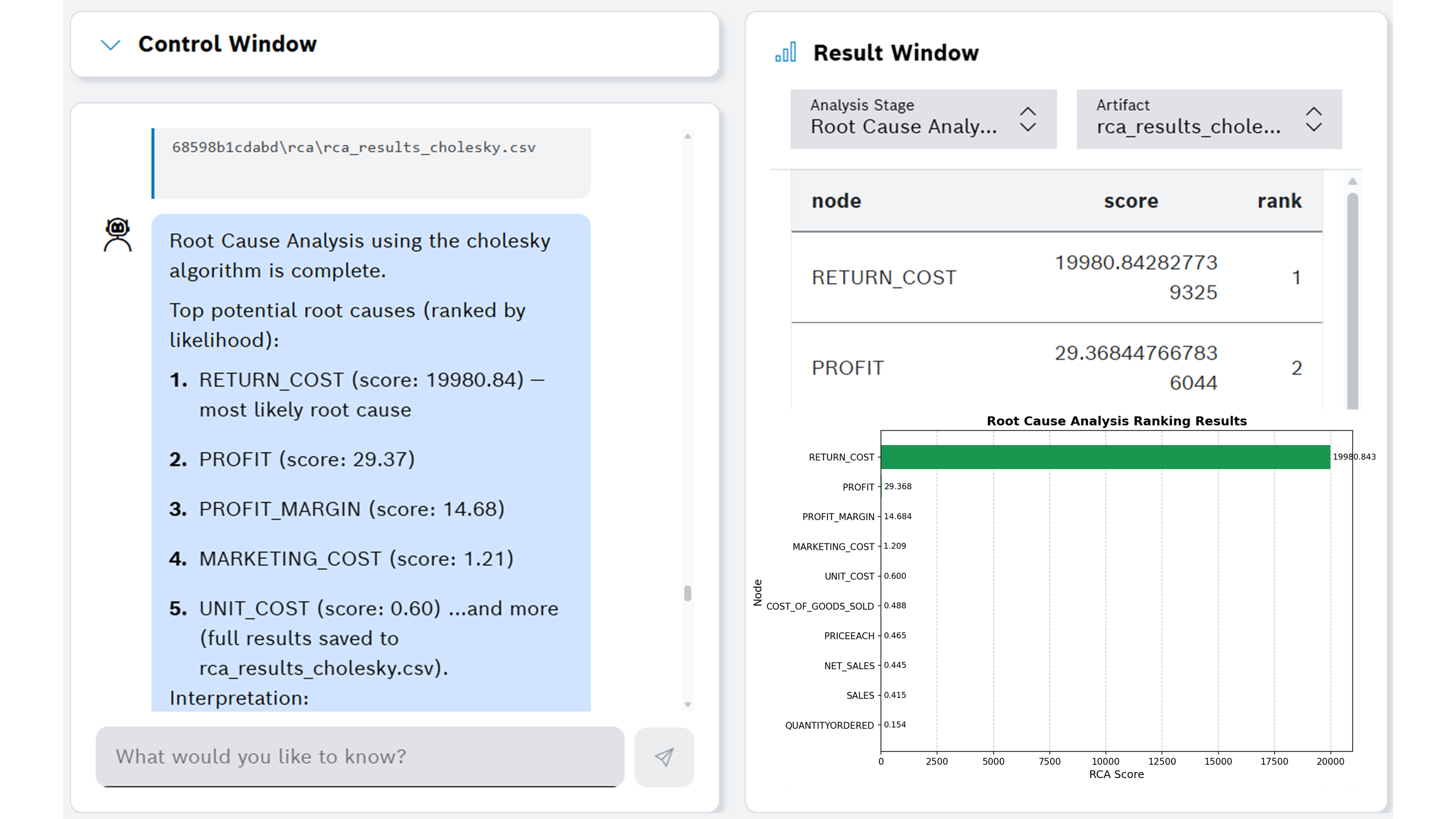}
        \caption{\small Retail}
        \label{fig:example_2}
    \end{subfigure}

    \captionsetup{width=\textwidth}
    \caption{\textbf{Scenarios}. Example of ORCA functionalities in different use-cases: a) Manufacturing, and b) Retail.}
    \label{fig:experiment2}

\end{figure*}

\section{Conclusion}

In this paper, we introduced ORCA, an LLM-powered interactive copilot designed to enable causal analysis. Through intuitive conversational interactions and a human-in-the-loop workflow, users can navigate complex tasks, including causal discovery and Root Cause Analysis (RCA). The system orchestrates the best pipeline by proactively suggesting the next steps, recommending optimal state-of-the-art algorithms based on dataset characteristics, and automating hyperparameter tuning. By abstracting the steep methodological complexity of causal inference, we believe ORCA represents a first-of-its-kind solution that empowers non-experts to independently extract causal insights from real-world data.

\section*{Declaration on Generative AI}
During the preparation of this work, the author(s) used Gemini in order to: perform stylistic editing, summarize text, and check grammar and spelling. After using this tool/service, the author(s) reviewed and edited the content as needed and take full responsibility for the publication’s content.

\bibliography{causal-copilot}

\appendix
\end{document}